# An Empirical Comparison of Algorithms for Aggregating Expert Predictions

**Varsha Dani**
Dept. of Computer Science
University of Chicago
Chicago, IL 60637

**Omid Madani and David Pennock**
Yahoo! Research Labs
74 N. Pasadena Ave., 3rd Floor
Pasadena, CA 91103

**Sumit Sanghai**
Dept. of Computer Science
University of Washington
Seattle, WA 98195-2350

**Brian Galebach**
ProbabilitySports.com

## Abstract

Predicting the outcomes of future events is a challenging problem for which a variety of solution methods have been explored and attempted. We present an empirical comparison of a variety of online and offline adaptive algorithms for aggregating experts' predictions of the outcomes of five years of US National Football League games (1319 games) using expert probability elicitations obtained from an Internet contest called ProbabilitySports. We find that it is difficult to improve over simple averaging of the predictions in terms of prediction accuracy, but that there is room for improvement in quadratic loss. Somewhat surprisingly, a Bayesian estimation algorithm which estimates the variance of each expert's prediction exhibits the most consistent superior performance over simple averaging among our collection of algorithms.

## 1 Introduction

Consider the problem of predicting outcomes of future events such as forcasting the weather, the stock markets, political races and sports games. In such prediction problems we often have access to extra information in the form of prediction probabilities, for the various possible outcomes, from a group of "experts". The task is then how to aggregate such information to more effectively predict future outcomes. A number of different methods have been studied and attempted for such prediction tasks, including information markets, polling, prediction methods based on machine learning, and belief aggregation methods [CBFD+97, SSWPG04, Kah04, GKV04, GZ86].

Unfortunately, often real data is not available and there is a lack of a comparison of various methods in actual settings. In this paper we describe a real domain, predicting US professional football games, using actual expert predictions from an online contest called Probability Sports [Pro]. We present a comparison of a number of algorithms for the task, ranging from the baseline of simple averaging of the expert predictions ("Average") to more sophisticated machine learning algorithms, including the experts algorithms of Cesa-Bianchi *et al.* [CBFD+97] and a novel algorithm which we refer to as the "Variance" algorithm. The Variance algorithm is a Bayesian estimation algorithm that models expert predictions as Gaussians centered around the actual outcome's probability, each expert's prediction having a different variance, and the algorithm attemps to estimate the variances from data seen so far.

We find that ProbabilitiySports is a challenging domain. For example, we observe that a variety of algorithms that we experimented with do not beat the baseline of simple averaging of experts' probabilities in terms of prediction accuracy (01 loss). These experiments provide some evidence that there is no room for improvement over Average in better predicting the binary outcomes, and thus the problem is purely about predicting better probabilities. Further, we observe that the simple Average algorithm is competitive with a variety of adaptive algorithms that we experimented with under the quadratic loss criterion. However, Cross-validation experiments and statistical significance tests over a few years worth of data indicate that some of our algorithms are superior to Average under the quadratic loss. In particular, the Variance algorithm appears to be the best in consistently beating Average in most experiments. As prediction problems are often difficult, small but statistically significant improvements in predictions can make a significant difference in the long run.

The paper is organized as follows. In Section 2 we define the problem and describe the data. Section 3 describes our methods, Section 4 presents our results together with analyses and discussions, and Section 5 discusses related work.

## 2 Problem Formulation

Our prediction task is to forecast the outcomes of a sequence of events. The events are binary and are represented as $y_1, y_2, \ldots, y_t, \ldots, y_T$ where $T$ is the total number of time steps and $t$ is a particular time point. The predictions of the experts for the event $y_t$ are represented as $p_t^1, p_t^2, \ldots, p_t^n$ where each $p_t^j$, $p_t^j \in [0, 1]$, is the probability that $y_t = 1$ according to the $j^{th}$ expert.

The prediction algorithm $A$ for each event $y_t$ takes as input the predictions $p_t^1, p_t^2, \ldots, p_t^n$ along with the outcomes $y_1, \ldots, y_{t-1}$ and the previous predictions $p_1^1, \ldots, p_1^n, \ldots, p_{t-1}^1, \ldots, p_{t-1}^n$, and outputs a probability $p_t^A$ that $y_t = 1$.

To measure the quality of the predictions made by the algorithm and the expert we shall use a loss function. An algorithm $A_1$ is supposed to perform better than $A_2$ under loss function $L$ iff $L(A_1) < L(A_2)$. The loss function for each event depends on the outcome $y$ and the predicted probability $p$ and it is denoted by $L(p, y)$. The total loss function is then the sum of the losses of the individual events, i.e., $L(A) = \sum_t L(p_t^A, y_t)$. Common loss functions include the absolute loss: $L(p, y) = |p - y|$, the quadratic loss: $L(p, y) = (p - y)^2$, and the log loss: $L(p, y) = -y \log p - (1 - y) \log(1 - p)$. Depending on the loss function an expert's strategy can vary. An important property of the quadratic and the log loss functions is that the experts' best strategy is to reveal their true beliefs [WM68]. We shall focus on the quadratic loss function for assessing the quality of probability predictions.

### 2.1 Data Set

The data that we worked with was collected from the Probability Sports website [Pro]. This is a game site that follows the National Football League (NFL) games and other leagues each year. Each season, participants make predictions (in the form of probabilities) on the outcomes of the games. On each round, based on the outcomes of the actual game, they receive a score according to a quadratic scoring rule. Their cumulative score for the season is the sum of their scores on individual games. As an incentive, at the end of each season the top few participants receive prizes. This has become a fairly popular site; the NFL competition in 2004 had 2231 participants.

The website uses a quadratic scoring rule which is given by $100 - 400(p - y)^2$ where $p$ is the predicted probability and $y \in \{0, 1\}$ is the outcome. The scoring rule is a scaled inverse quadratic loss function described earlier. Thus, when experts are trying to maximize their scores, their best strategy is to give their true beliefs. This makes this data close to an ideal testbed for our experiments[1]. We treated the predictions of the participants in the game as expert advice for our prediction algorithms. We use no additional information like the previous records of the NFL teams participating in the game.

Data is available for the football games from 2000 onward. The total number of games for all 5 seasons is 1319, just over 250 per season. Figure 1 shows the ranked (final) scores of the experts. Median of the final scores for the years 2000 through 2004 were -485, -649, -684.2, -437, and -275 respectively, while the averages were -1301, -1547, -1792, -1221, and -944 respectively.

## 3 Algorithms

### 3.1 The Experts Algorithm

The basic premise of the experts algorithm of Cesa-Bianchi *et al.* is to predict according to the weighted average of the experts' advice. The weights associated with the experts are changed dynamically based on their ongoing performance. We briefly state the algorithm here for completeness.

```
0  Given a parameter β
1  Start all experts with equal weights
2  For each trial:
3      r = weighted average of expert predictions.
4      Predict p = F_β(r)
5      Observe the true outcome y.
6      Update each expert's weight
```

The function $F_\beta$ on line 4 is called the prediction function. Any function may be used as the prediction function, provided it satisfies the bounds

$$1 + \frac{\ln((1-r)\beta + r)}{2\ln(2/(1+\beta))} \leq F_\beta(r) \leq \frac{-\ln(1 - r + r\beta)}{2\ln(2/(1+\beta))}$$

In fact it is not even necessary to use the same prediction function on every round. The guarantees of the experts algorithm hold as long as the predictions made by the algorithm are always within the range specified by the above bounds. The weights are updated according a multiplicative update rule; the new weight of an expert with weight $w$ whose prediction is off from the true (binary) value by $q$ is $wU_\beta(q)$. Here again any update function $U_\beta(q)$ may be used provided it satisfies the bounds $\beta^q \leq U_\beta(q) \leq 1 - (1 - \beta)q$. The weights are a measure of the relative credibility of the experts.

Cesa-Bianchi *et al.* showed that the loss of the algorithm is bounded by $\frac{\ln N + L \ln(1/\beta)}{2\ln(2/(1+\beta))}$, where $L$ is the loss of the best expert (in hindsight), $N$ is the number of experts and $0 < \beta < 1$ is a parameter used by algorithm. Note that this is an *a posteriori* guarantee on the loss of the algorithm. We do *not* need to know who the best expert will be during the run of the algorithm.

---

[1] As this is a contest where only the top few win prizes, the incentive structure is not exactly truth revealing.

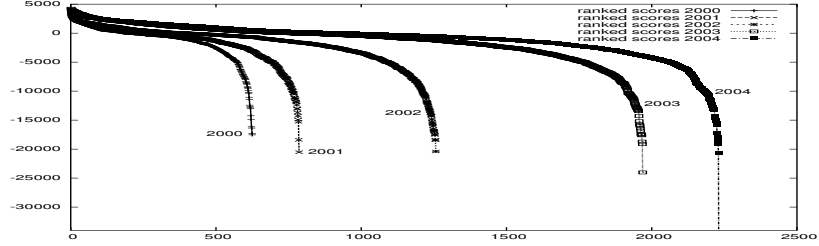

Figure 1: Ranked final scores of the experts in 5 seasons of NFL data.

We experimented with a number of variants of this algorithm, based on the prediction and update functions used and the method of handling data. We tried three different prediction functions: (1) Vovk's function [Vov90], $F_\beta(r) = \frac{\ln(1-r+r\beta)}{\ln(1-r+r\beta)+\ln((1-r)\beta+r)}$, (2) A piecewise linear function that satisfies the bounds specified by the analysis of the experts algorithm: $F_\beta(r) = 0$ if $r \leq \frac{1}{2} - c$, $F_\beta(r) = \frac{1}{2} - \frac{1-2r}{4c}$ if $\frac{1}{2} - c \leq r \leq \frac{1}{2} + c$, and $F_\beta(r) = 1$ if $r \geq \frac{1}{2} + c$, where $c = \frac{(1+\beta)\ln(2/(1+\beta))}{2(1-\beta)}$, and (3) $F_\beta(r) = r$ (This identity function does *not* satisfy the bounds specified in the analysis, but seems to work reasonably well nevertheless).

We tried three different update functions $U_\beta(q)$: (1) $U_\beta(q) = e^{q\ln(\beta)}$, (2) $U_\beta(q) = e^{-\beta q}$, and (3) $U_\beta(q) = 1-(1-\beta)q$. In each case the old weights are multiplied by $U_\beta(q)$ where $q$ is the loss incurred on the given round.

The algorithm as specified assumes that every expert provides advice on every round of play. Our dataset, on the other hand, contained missing data, where some of the experts did not provide any predictions on some rounds. One way to handle this problem is to treat a missing prediction as a prediction of 50% (the event occurs with probability $\frac{1}{2}$). Note that this is the prediction that minimizes the loss averaged over both binary outcomes. Another possibility (assuming, not unreasonably, that on every round there is *some* expert who offers advice) is to only take the experts who did offer advice into consideration when making a prediction. While updating the weights, one can then modify the weights of the experts who didn't participate on that round so that their *relative* weight stays the same.

### 3.2 The Variance Algorithm

In the experiments section we will see that a simple averaging of the experts' prediction is competitive with the experts' algorithm. If we make the assumption that the experts' predictions are samples of a Gaussian distribution centered around the true probability of the event (one Gaussian per expert), then the mean of the experts' predictions will converge to the true probability as we increase the number of experts. However, not all experts behave in the same fashion. For example, some experts are more aggressive than others, some are more informed than others, etc. We will try to capture this notion by assuming that the Gaussian for expert $i$ has variance $\sigma_i^2$ (which we assume stays the same across all events). Each event $y_t$ is associated with a "true" probability $p_t$ from which the outcome is drawn. An expert's prediction for event $y_t$ is then assumed to be drawn from her Gaussian distribution centered around $p_t$ with variance $\sigma_i^2$.

Since one wants to minimize the quadratic loss function, the predictor's best strategy is to predict the true probability, in this case $p_t$. If we knew the variance of all the experts, the true probability can then be computed as the one that maximizes the likelihood of the observed experts predictions. We know that $p_t^i \propto e^{-(p_t^i-p_t)^2/\sigma_i^2}$. Assuming independence between the experts' predictions, maximum likelihood $p_t$ is the one that maximizes $\prod_i e^{-(p_t^i-p_t)^2/\sigma_i^2}$ and is given by

$$p_t = \frac{\sum_i w_i p_t^i}{\sum_i w_i} \quad (1)$$

where $w_i = \frac{1}{\sigma_i^2}$. On the other hand if we knew the true probabilities along with the expert's prediction of all the events, the variance of the expert can be estimated as

$$\sigma_i = \sqrt{\frac{\sum_t (p_t - p_t^i)^2}{T}} \quad (2)$$

Given just the predictions of the experts, calculating the true probabilities and the variances is an optimization. We use an EM-like iterative approach [DLR77] where we start with equal variances for all the experts and use Equation 1 to get the true probabilities. Now given the true probabilities of all events we compute the variances using Equation 2 and repeat this procedure. To reduce the time complexity of this algorithm, for each event we initialize the variances to the ones calculated for the previous event. Thus, starting the procedure with some *a priori* estimates of the experts' variances, we can alternate these computations to get better and better estimates of the variances and true probabilities.

Although the variance approximation is hinged on some simplifying assumptions, in the experimental section we see that the method works well in practice.

### 3.3 Other Machine Learning Approaches

An experts algorithm may be viewed as an online and thus a single pass machine learning algorithm: it adapts the experts weights by "touching" each instance once (hence a single pass over the training data). We also experimented with a variety of other "batch" learning algorithms that are not necessarily single pass, such as multi-pass versions of perceptron and winnow algorithms that attempt to minimize an objective function such as 01 error (maximize prediction accuracy), linear and nonlinear support vector machines (we tuned the regularization constants), as well as other algorithms such as decision trees and ensemble methods (bagging and boosting over decision trees and decision stumps). We evaluated the methods under both prediction accuracy performance and quadratic loss performance, and under both the online setting (train on the instances seen so far, and test on the next instance), and in the cross validation setting (*i.e.*, ignore the order of instances, and average performance on a large number of train and test splits of the data). Unfortunately, none of the standard classification methods performed better than simple averaging, in either accuracy or quadratic loss (*e.g.*, see Section 4.1 for linear SVM results). Here, we will describe a batch learning algorithm, *exponentiated gradient* (Exp Gradient), based on an exponential updating technique [KW97], which performed competitive with other experts algorithms and appears to beat simple averaging in cross-validation experiments.

Exp Gradient is designed to minimize quadratic loss (via the exponentiated gradient). In every pass over the training data, the algorithm updates the expert weights on *every instance*, using the following weight update formula:

$$w_i \leftarrow w_i * exp(2.0 * x_i * \delta * l_r),$$

where $x_i$ is the probability given by expert (feature) $i$ in the given instance (game), $\delta$ is the difference $y - p$, between the desired output $y$ (1 or 0 depending on the outcome of the game) and output $p$ of the current predictor, and $l_r$ is the learning rate. The weights are then renormalized to sum to 1, and the next instance is examined. A missing value is treated as a 0.5 prediction.

There are a number of variations to the basic algorithm, *e.g.*, whether to randomize orderings of instances and whether to continue the passes until error is no longer lowered or whether to stop after a fix number of passes, and the choice of the parameter values to be set (*e.g.*, learning rates and number of passes). We experimented with a few variations on the NFL seasons 2000 through 2003, and found the following settings to perform satisfactory: We set the number of passes to 3 (more passes often resulted in inferior performance possibly due to overfitting). The learning rate is fixed at $l_r = 0.1$. Ideally, these parameters could be set dynamically via extensive cross-validation. The instances are visited in chronological order. At the end of each pass, the error (quadratic loss on training data) for the current weights (predictor) is computed, and the best predictor across the different passes is selected to predict the probability for the next game. Randomizing the initial weights of experts or the order the instances are visited were found to have insignificant effects or result in inferior performance. Interestingly, forming a committee by training over bootstrap samples of the training data, to adjust for possible overfitting (to achieve smoothing), did not appear to help. We also experimented with variations of additive rather than exponential updates, but that resulted in inferior performance (even after normalizing the feature weights so outputs of the predictor during training lies in 0 and 1). This is possibly due to some expert (feature) weights becoming negative in standard additive updates: the search space of unmodified additive update algorithms, which allows negative weights, may not be appropriate. This may explain the inferior performance of SVMs as well (see Section 4.1). We note that there are variants of additive updates that ensure that the expert weights remain non-negative [KW97].

### 3.4 Market Simulation

Prediction markets and betting markets have been shown to effectively aggregate the opinions of traders and provide accurate forecasts. We implement a simulated information market, where agents in the market correspond to experts. Each agent has a prior belief equal to the prediction given by the expert, and logarithmic utility for money. For each game, agents buy and sell a security paying off $1 if and only if Team A wins. Agents reach an competitive equilibrium where the supply from high-belief agents meets the demand from low-belief agents. Moreover, we simulate agents learning from the market by setting their posterior belief in equilibrium to be the average of their prior be-

lief and the equilibrium price [PW01]. Agents gain or lose money in each round; agents that are more accurate will tend to gain money. The additional money is reflected back into their utility function, and agents with more money will tend to risk more money and thus have higher weight in the future. The aggregate prediction is taken to be the equilibrium price in the market, which in the case of logarithmic utility happens to be a form of a wealth-weighted average [PW97, PW01]. Since more accurate agents tend to accrue more wealth, the market simulation can be thought of as a variant of the expert algorithm. Note however that the market simulation does not satisfy the worst-case bounds of other expert algorithms. If an agent has belief 0 or 1, she will bet all of her wealth and, if incorrect, will be eliminated from all future rounds. Clearly an adversary would first eliminate some expert in round 1, then make that expert the only informative predictor after round 1. Even though the market has no worst-case guarantee, it still can make for a reasonable expert algorithm in terms of average case behavior.

## 4 Experiments

### 4.1 Performance on Accuracy

One approach to obtaining better probability outputs is to first seek learning algorithms that do well on the accuracy objective, *i.e.*, in minimizing the (zero-one) misclassification rate, and then to extract probabilities from those algorithms, for example via subsampling [LZ05]. As there has been much work on obtaining robust classification algorithms, this is an attractive approach. However, both in the online and in cross-validation experiments, we observed that there was no evidence that any of the machine learning algorithms we tested (*e.g.*, decision trees with bagged and boosted variants, SVMs), as well some of the probability prediction algorithms reported in this paper (some of which tend to perform better in outputting probabilities), consistently did better in zero-one error performance over simple linear averaging (Average). (For each game, we averaged predictions only over experts that provided a prediction for that game.) Figure 2(a) presents the zero-one errors for the first four years for several methods. For linear SVMs [CL01], the numbers correspond to (3-fold) cross-validation, where we used the best parameter settings observed on held-out data (a regularization constant of 1, and the data was L2 normalized). Note that the zero-one errors for Average and the best expert would be the same whether we use cross-validation or measure the online variant, as simple averaging and top expert ignore the "training data". The standard deviation (over the different held-out data samples) for simple averaging and top expert was about 0.05 for the different scoring years.

These validation experiments revealed that many learning algorithms (at least when not significantly altered) were inferior for this task, under either accuracy or quadratic loss. This is the case even when given a fair amount of training instances as in cross-validation experiments (as opposed to the online setting in which the algorithm has to begin predicting with little data). We do not report on their performance in the online experiments, except for the Exp Gradient algorithm. Furthermore, these experiments suggested that in terms of accuracy, there is not much to be gained beyond simple averaging: it's not even clear that the top (scoring) expert, who decidedly beats Average under the quadratic scoring criterion (equivalently, the quadratic loss), beats Average in 01 error (Figure 2(a)).

On the other hand, cross-validation experiments also suggested that some algorithms (*e.g.*, the average of top $k$ experts on the training set, with say $k = 30$) can beat simple averaging under the quadratic scoring criterion. Average obtains a score of around 3000 per season (see 4.2). To get an idea of the room for improvement, note for example that even if a method has a 01 error of 40% (worse than the error of Average), if it is almost perfect at outputting probabilities, its score in a 250 game season could reach 15000: it could obtain close to 100 points on games it predicts accurately by outputting a confident probability (0 or 1). This is roughly 60% or 150 many of games, and on the remaining 40% it could output a probability close to 0.5. Also note that a conservative method, which would output a probability according to the roughly 0.35 error rate of Average, *i.e.*, for each event, it would output a probability of 0.65 whenever the probability given by Average was above 0.5 (*i.e.*, Average predicted a win for the outcome), and 0.35 otherwise, would underperform and obtain a score of roughly 2250 for the same period.

### 4.2 Performances on Quadratic Score

Fig. 3(a) shows the scores obtained by the various algorithms and the score of the top expert. All algorithms except for Average are *adaptive*, *i.e.*, they use their training set, the games played so far, to adjust the experts' weights. Average and Variance algorithms did not require any parameters. For the remaining algorithms, we used the years 2000 to 2003 for selection of good parameter values (*e.g.*, number of passes for Exp Gradient, or the update function for the experts algorithm). The number 30 in "Average (30)" refers to the variant in which at each time point only the top 30 scoring experts at that point are used to give the prediciton for the next game. Similarly, for Variance

| Year | 2000 | 2001 | 2002 | 2003 | 2004 |
|---|---|---|---|---|---|
| No. of experts | 625 | 786 | 1257 | 1969 | 2231 |
| Top Expert | 3185 | 3445 | 3339 | 4218 | 3747 |
| Average | 2561 | 2574 | 2562 | 3298 | 3371 |
| Average (30) | 2864 | 2589 | 2529 | 3731 | 2986 |
| Variance | 2979 | 2660 | 2627 | 3498 | 3456 |
| Variance (20) | 3187 | 2662 | 2611 | 3881 | 3344 |
| Experts | 2801 | 2541 | 2406 | 3343 | 3099 |
| Expert MD | 2875 | 2644 | 2505 | 3442 | 3346 |
| Exp Gradient | 2827 | 2563 | 2616 | 3371 | 3137 |
| Market Sim | 3090 | 2482 | 2381 | 3397 | 3203 |

(b) NCAA

| Year | 2001 | 2002 | 2003 |
|---|---|---|---|
| Top Expert | 2251 | 2401 | 2487 |
| Average | 1804 | 1607 | 1631 |
| Experts | 1827 | 1595 | 1643 |
| Variance | 1844 | 1706 | 1719 |

Figure 3: (a) Scores on the NFL data for different years. Expert MD refers to the missing data variant of the experts algorithm. Average (simple averageing) is competitive with the adaptive adaptive algorithms, but the Variance algorithm consistently outperforms it. (b) Scores on the NCAA data.

(a)

| Year | 2000 | 2001 | 2002 | 2003 |
|---|---|---|---|---|
| Home Team | 0.4363 | 0.4479 | 0.412 | 0.3895 |
| Top Expert | 0.3514 | 0.3127 | 0.3521 | 0.3221 |
| Average | 0.3552 | 0.3436 | 0.3708 | 0.3109 |
| SVM | 0.3808 | 0.3571 | 0.3826 | 0.3179 |

(b)

| Period | 2000-3 | 2001-3 | 2002-3 |
|---|---|---|---|
| Top Expert | 9910 | 8533 | 6782 |
| Average | 11169 | 8867 | 6221 |
| Variance | 11512 | 8774 | 6120 |

Figure 2: (a) Prediction errors (zero-one) for predicting a win for the home team, the top expert, and Average and linear SVM methods. (b) Scores in the multi-year experiments.

(20) algorithm, the top 20 experts are chosen dynamically to be the ones that have the least variance up to that point in time.

The first and foremost observation is that the top expert beats all the algorihms, but her score is not much greater than the scores obtained by the various strategies. It is also worth noting that in most years the algorithms come within the top 20 positions at the end of the season. Secondly, Average performs considerably well compared to the adaptive algorithms, considering its simplicity. Furthermore, one sees that while in the first season (2000), all the adaptive algorithms beat Average fairly significantly, this gap goes down with subsequent years, even though the number of participating experts increases every year. This perhaps implies that the experts population is getting better or more competitive as a whole (so Average becomes competitive).

As the average and the median of the final scores of experts are negative (Section 2.1), the question arises as to why Average does relatively well. The reason for the poor individual scores, but good Average scores lies in the fact that most experts are not well-calibrated, i.e., while they may often predict well, each such expert gives extreme probabilties in sufficiently many games and looses too many points to end up with a poor final score. However, even the simple average of only the experts that ended up with a score below 0 at the end of each season, yields scores of 1763, 2074, 2051, 2697, and 2717 for the years 2000 through 2004 respectively. These positive scores rank fairly high in the corresponding seasons (respectively, 33, 23, 34, 78, and 62). This implies that on average, for every game, the "good" predictions outweigh the "bad" ones. Thus the averaging procedure is effective in smoothing the extreme probabilities and turning the individual negative final scores into competitive overall positive population scores.

Within the variants of the experts algorithm, we use the update function $U_\beta(q) = e^{-\beta q}$, with $\beta = 0.75$, which appeared superior to the others. For prediction functions, Vovk's function and the piecewise linear one seem to have comparable performance, and we use Vovk's function. A significant advantage is obtained by using the missing data variant of the experts algorithm, where the relative weight of an expert is only changed on rounds in which they participate. However, none of the experts algorithm is consistently better than Average in every season. In many cases this can be attributed to the fact that many aggressive experts (that are also predicting accurately) receive high weights in the initial games and following these experts in the future rounds results in more losses than gains.

The variance algorithm seems to have good performance all around. One can also conclude that the variance algorithm consistently outperforms Average. We conducted a sign test on the years 2003 and 2004 as follows: for every game, we recorded whether or not Variance gets a higher score (lower quadratic loss) than Average, and found that the number of wins of Variance was higher than its losses at significance level $p <= 0.1$ in both years. Thus we believe that, from the algorithms we tested in this domain, Variance is the most consistent best performing algorithm in sin-gle season periods.

Since the number of experts is considerably more than the total number of games, the algorithms may not have enough data to train on. Hence, we ran the algorithms on multi-year periods, in which we consider only those experts that played in all the years of that period. Along with increasing the number of events, a side-effect is that the number of experts is reduced to around 100. Figure 2(b) shows the results of our experiments, for periods 2000-3 (years 2000 through 2003), 2001-3, and 2002-3. We see that the Variance and Average algorithms beat the top scoring expert (for that period) in two of the three cases (the two longer time periods), confirming our intuition.

In 2003, two of the participants in the game were Tradesports [Tra] and Newsfutures [New]. These are online information market sites where people trade on securities for events. There is widespread and compelling empirical evidence that the prices of securities in such markets are good estimators for the probabilities of the events that they represent [WZ04]. In the 2003 season these markets were introduced as participants on Probability Sports, for a study by Servan-Schrieber *et al.* [SSWPG04]. The prices of the securities for the games in the markets were entered as predictions in the game. Both markets finished close to the top (although they were ineligible for prizes) coming in the sixth and eighth places with scores of 3389 and 3359, respectively. The Variance algorithm's score is competitive with the markets, but results from additional years are needed for a better comparison.

We also applied some of our algorithms to the NCAA basketball data available on the same site. The data only includes the playoff games and thus the number of games is much less (roughly 60). In this case, the variance algorithm also gives the best results (Fig. 3(b)). All in all, Variance beats Average in 9 out of our 11 experiments (5 NFL and 3 NCAA seasons, and 1 out of 3 multi-year experiments).

### 4.3 Discussion

The Variance algorithm was motivated by the obervation that simple averaging of expert predictions is competitive with other algorithms,[2] and was based on the core assumption that every expert's prediction is a sample from a distribution centered around the "true probability" of an event. We made a number of additional simplifying assumptions in its derivation, for example that the experts distributions are Gaussians that do not change with time, and that the experts are independent. Lifting some of the assumptions may lead to superior performance for the Variance algo-rithm. Perhaps, the most peculiar aspect of the Variance algorithm is that it does not take the outcomes of the previous games and experts' performances into account (directly). Yet, Variance is the most successful algorithm that we tested in this domain.

---

[2]In our experiments, on the years 2000 through 2003.

The Variance algorithm does not require any parameters, while others such as the experts algorithms and Exp. Gradient require a number of choices to be made. However, the experts algorithms often carry worst-case guarantees and are fairly versatile. Obviously, the success of Variance depends heavily on the accuracy of its core assumption. It is possible that an experts algorithm on top of several baseline aggregation algorithms (Average, Variance, Average (30), ..) may lead to improved or more reliable performance.

## 5  Related Work

An opinion pool is a mathematical aggregation function for combining expert beliefs, for example a weighted algebraic or geometric average. Opinion pools are usually justified on an axiomatic basis [GZ86]. Computational expert algorithms on the other hand are typically evaluated according to worst-case performance. A few studies of belief aggregation have empirical components. Ng and Abramson [NA92] simulate the distribution of experts' beliefs as a Guassian around the "true" probability and test several opinion pools, concluding that a weighted average works best. Three recent papers leverage the ProbabilitySports data. Servan-Schreiber et al. [SSWPG04] compare the accuracy of real-money and play-money prediction markets as additional "contestants" in the 2003 ProbabilitySports contest. Wolfers and Zitzewitz [WZ05] derive a number of models where prediction market prices can be interpreted as functions of expert beliefs, and show that the ProbabilitySports data fits their model quite well. Chen et al. [CCMP05] take an approach most similar to ours, comparing different belief aggregation methods on the 2003 data. As their focus was to compare against prediction markets and had market data only for 2003, their study is limited to 2003 data. They tested weighted algebraic and geometric averages of various types; we have examined a wider variety of machine learning methods. Their conclusion was that a simple average was about as accurate as prediction market prices. In this paper, we find evidence that, although it is very hard to beat the average, one algorithm (Variance) consistently outperforms the average.

## 6  Conclusion

Our experiments with a variety of algorithms and the difficulty of beating the baseline suggest that the probability prediction problem in Probability Sports, be-yond that achievable by the baseline of simple averaging, is a difficult task. However, our experiments provide evidence that the Variance algorithm yields the most consistent superior performance compared to the other algorithms.

Future directions include extensions to the Variance algorithm, for example in estimating experts' biases in addition to the variances, dropping the independence assumption, and weighting the recent history more, as well as developing a better understanding for reasons for its superior performance and the validity of its core assumption. We would also like to experiment with other algorithmic ideas and in particular with algorithms that directly attempt to minimize quadratic loss on the training data using appropriate regularization constraints. Taking other contextual information into account, such as the division and the identity of the teams playing the game (some experts may excel at predicting certain divisional games), may also boost performance.

ProbabilitySports is a realistic and challenging domain for evaluation of prediction algorithms, and in particular for comparison of algorithms for aggregating expert predictions. The data accumulated in the upcoming years can only add to value of the collection. We plan to make the current data available in the UCI repository [BKM98].